# Modified Bat Algorithm: A Newly Proposed Approach for Solving Complex and Real-World Problems

Shahla U. Umar[1], Tarik A. Rashid[2], Aram M. Ahmed[2], Bryar A. Hassan[2,3], Mohammed Rashad Baker[1]

[1]Network Department, College of Computer Science and Information Technology, Kirkuk University, Kirkuk, Iraq.

[2]Computer Science and Engineering Department, School of Science and Engineering, University of Kurdistan Hewler, KRI, Iraq.

[3]Department of Computer Science, College of Science, Charmo University, 46023 Chamchamal, Sulaimani, Iraq

Email (Corresponding): shahla_umar@uokirkuk.edu.iq

**Abstract**

Bat Algorithm (BA) is a nature-inspired metaheuristic search algorithm designed to efficiently explore complex problem spaces and find near-optimal solutions. The algorithm is inspired by the echolocation behavior of bats, which acts as a signal system to estimate the distance and hunt prey. Although the BA has proven effective for various optimization problems, it exhibits limited exploration ability and susceptibility to local optima. The algorithm updates velocities and positions based on the current global best solution, causing all agents to converge towards a specific location, potentially leading to local optima issues in optimization problems. On this premise, this paper proposes the Modified Bat Algorithm (MBA) as an enhancement to address the local optima limitation observed in the original BA. MBA incorporates the frequency and velocity of the current best solution, enhancing convergence speed to the optimal solution and preventing local optima entrapment. While the original BA faces diversity issues, both the original BA and MBA are introduced. To assess MBA's performance, three sets of test functions (classical benchmark functions, CEC2005, and CEC2019) are employed, with results compared to those of the original BA, Particle Swarm Optimization (PSO), Genetic Algorithm (GA), and Dragonfly Algorithm (DA). The outcomes demonstrate the MBA's significant superiority over other algorithms. Additionally, MBA successfully addresses a real-world assignment problem (call center problem), traditionally solved using linear programming methods, with satisfactory results.

**Keywords**

Bat Algorithm, Modified Bat Algorithm, Bio-inspired Algorithm, Swarm Intelligence, Control Process.

## 1. Introduction and Related Work

Traditional search algorithms are efficient in several fields; however, they have their limitations. For example, most of them are deterministic, and the same output will be obtained for any given input. Also, for optimization problems, there is no guarantee to reach the global optimum. Additionally, they tend to be suitable for problem-specific. Subsequently, heuristic and meta-heuristic algorithms based on the trial and error concept were developed to minimize the mentioned limitations and to obtain an optimal solution in a reasonable time [1]. Similarly, these algorithms try to apply randomization techniques and local search in different methods [2][3]. Heuristic algorithms have been developed to make what is known as meta-heuristic algorithms, which have better performance than heuristic algorithms [4]. Over several years, intelligent behaviors and biological actions with adaptability and self-learning ability have appeared in biological species (fish, bats, insects, and birds), which led to the so-called swarm intelligence. To understand how these animals, solve their problems, many researchers have studied these behaviors and natural phenomena. Swarm agents reinforced the exploration mechanism in the complex search space obeying some rules and instructions without any central control. Thus, in an attempt to imitate the inherent features of such biological systems for handling various simulation and optimization problems, different researchers in this field investigated computational models and have simulated bio-inspired intelligent behaviors in the form of computational algorithms.

The beginning of swarm intelligence algorithms was in the 1960s at the University of Michigan, in 1960, John Holland and his colleagues wrote their first book on the GA, and the development of their book was published in 1970 and 1983 [5]. In the same year, a new algorithm was developed, which is based on the annealing process of the metals, known as simulated annealing (SA) [6]. Also, there has been a remarkable development in algorithms inspired by nature, for example, in 1995 James Kennedy and Russel C. Eberhart proposed PSO, this algorithm was inspired by the natural intelligence of a swarm of fish and birds [7], later PSO became the foundation for several other algorithms, such as DA. A vector-based differential evolution algorithm was proposed in 1997, which outperformed GA in different applications [8]. For various optimization problems like transport modeling and water distribution, a new algorithm was presented in 2001 called Harmony search (HS) [9]. For internet hosting center optimization problems, in 2004, a new algorithm known as the Honey Bee Algorithm (HBA) [10]. After that, [11] proposed the Artificial Bee Colony (ABC), and in 2009 and 2010, XinShe Yang developed the Firefly Algorithm (FA) [12] and the Cuckoo Search Algorithm (CS) [13] respectively. Also, in 2010, the same author proposed the BA [2]. In 2015, a new algorithm was developed, which is based on the PSO algorithm inspired by the behaviors of the dragonfly swarm of attracting the enemy to the food, which is called DA [14]. The same author proposed two other algorithms, the Whale Optimization Algorithm (WOA) in 2016 [15] and the Salp Swarm Algorithm (SSA) in 2017 [16]. Although the novel ABC has a good exploration ability, it suffers from low exploitation, so to enhance this feature two new types of ABC were suggested in 2017. In the first algorithm, they developed an adaptive approach for the population size (AMPS) [17], while a ranking-based adaptive ABC algorithm (ARABC) [18].

BA Modifications, including EBA, MBA, Nonlinear optimization, HAM, and PCA-BA, have been conducted on the Bat Algorithm. To overcome the flaws in BA, [19] suggested a new algorithm called (EBA) that improves the exploration ability by making the value of the loudness (A) and the pulse rate (r) equal to the number of problem dimensions. The experimental results on 20 benchmark functions with different dimensions proved the efficiency of EBA against BA. In 2015, [20] proposed an MBA algorithm as an expert planner for sports coaching sessions, the results showed the predicted plan complied with the high standards of the cycling coach. In 2016, [21] proposed two modifications to the Bat Algorithm. They modified the acceptance scheme to reduce the probability of acceptance of worse solutions, and they changed the velocity update equation by archiving components and introducing the cognitive coefficient, as the suggested modification was done on the linear velocity equation, it did not affect the computational complexity or the fitness function evaluation of the algorithm.

In 2018, [22] modified the BA algorithm using the ABC to address the local optima problem. Their algorithm consists of two modes components. For this, they used a mutation factor and the technique of point generations. Then, the ABC structure is used to enhance the local search ability. Two new variations of principle component analysis (PCA-BA and PCA-LBA) were proposed in 2019 [23] to evaluate and enhance the global search capability of BA with large-scale challenges. The effectiveness of this new tactic is increased by determining a correlation threshold and generation threshold utilizing the golden section method.

The main objective of this paper is to improve the standard BAT algorithm. Hence, this research focuses on different aspects such as a brief explanation of the standard Bat algorithm, its limitations, and how it can be enhanced.

The main contribution of this paper is as follows: the new MBA is presented to improve the outcome of the existing BA as it is similar to the PSO [24] in terms of updating agents' positions ($x_i$) and velocities ($v_i$) and the parameters of updating the Bat algorithm equations essentially would control the convergence rate and the directions of the agents in the search space, consequently, the MBA is applied to the enhance the global search ability of the standard BA through adjusting these equations to include the frequency and the velocity of the best solution found so far as well as the value of the current global best solution.

The proposed MBA proves its ability to overcome local optimum trapping problems and increases the convergence rate by directing the search agents toward the most promising regions. The statistical outperforming of the MBA is proved by using different benchmark test functions (classical benchmark functions, CEC2005, and CEC2019) with remarkable comparative results on others. Finally, the case study aims to optimize the inbound call handle time for a call center, which receives about two hundred thousand calls a day. The problem is that there is a time mismatch between the time required to answer calls and the available time for agents, therefore, the modified Bat algorithm is managed to optimize this problem and minimize the average call time with very satisfactory results.

The rest of the paper highlights the following: description of the BA (its ability and features) in Section 2, while Section 3 clarifies the newly suggested modified BA. The implementation and the obtained results are discussed in Section 4. The comparison with the original BA is presented in Section 5, while Section 6 compares with the most common metaheuristic algorithms. Section 7 explains how to use the suggested algorithm to solve real-world Business Process Optimization problems. And finally, the conclusion is in Section 8.

**2. Bat Algorithm**

BA is a Meta-heuristic algorithm introduced by Yang in 2010. This algorithm depends on the echolocation ability of microbats directing them on their hunting behavior [2]. Bats are a wonderful group of animals, they are considered unique mammals that have wings and also can determine the location of prey by echo, and the statistics indicate the existence of hundreds of different species, which may reach 20% of the total mammals in the world.

## 2.1. Echolocation Capability of Bats

Ultrasonic echolocation signals in frequency usually range from 20 to 200 kilohertz (kHz), while the human ear may normally hear up to about 20 kHz. However, we may hear a sound echo shot from some types of bats. These bats beam loud sound pulses and listen to the echo from the surrounding objects, as shown in Figure 1. Bats emit pulses that differ in attributes depending on the prey-hunting approach and the species. So, this exciting behavior of microbats can be exploited in some manner that it can be used to optimize the objective function of different optimization problems and simulate bats' echolocation strategies to formulate novel optimization algorithms [25].

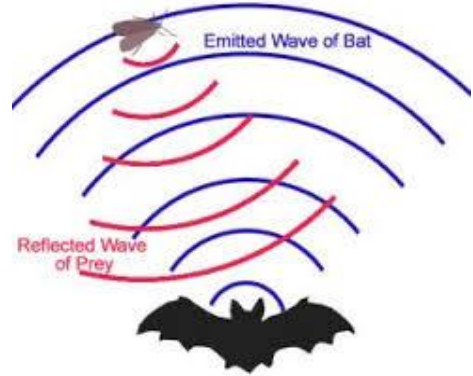

Figure 1: The bat's echolocation [25].

In the bat algorithm, the characteristics of the echolocation behavior of microbats can be exploited to improve different bat-inspired algorithms by benefitting from these features of bats [26].

To sense distance, all bats use echolocation and also, and they expertly distinguish between prey and obstacles. Bats fly randomly with velocity ($v_i$) at position ($x$) with a fixed frequency ($f_{min}$), varying wavelength ($\lambda$), and loudness ($A_0$) to search for prey. They can automatically adjust the wavelength (or frequency) of their emitted pulses and adjust the rate of pulse emission $r \in [0, 1]$, depending on the proximity of their target; although the loudness can vary in many ways, we assume that the loudness varies from a large (positive $A_0$) to a minimum constant value ($A_{min}$).

For optimization problems with *n*-dimensional search space, each bat is considered a solution that can be evaluated with the problem fitness function. Each bat in the population has two real-valued vectors, the first one represents the position of the bat in the problem search space, while the second vector clarifies the velocity of bats in (*n*) directions. Often, at the initial step of the algorithm, these two vectors are assigned randomly. At each iteration step, a new fitness value is calculated for every bat in the population, and the velocity, as well as the position vectors, are also updated according to the following equations: (1), (2), and (3) [21].

$$f_i = f_{min} + (f_{max} - f_{min})\beta \qquad (1)$$

$$v_i^{t+1} = v_i^t + (x_i^t - x_*)f_i \qquad (2)$$

$$x_i^{t+1} = x_i^t + v_i^{t+1} \qquad (3)$$

Where $x_i$ refers to the location of bats in the search space, while ($v_i$) expresses the speed of bats in that space and ($f_i$) is used to denote the frequency of the waves, whereas ($ß$) represents a vector of numbers selected randomly ranging between 0 and 1 and from determining distribution, also ($x_*$) is the best solution among the bats obtained to that moment. The highest value (upper limits) and the smallest value (lower limits) are determined depending on the dimensions of the specific optimization problem. In the beginning, a random value of the frequency is assigned to each bat ranging between ($f_{min}$, $f_{max}$). During the search for prey, as soon as the bat hunts their prey, the proportion of the loudness decreases, while the proportion of the pulse emission increases. Here is the pseudo-code invented by Yang [27].

**Algorithm (1): The Pseudo-code of the Bat algorithm**

Objective Function $f(x)$, $x = (x_1, \ldots x_d)^T$
Initialize the search agents (bat individuals) $x_i$, $(i = 1, 2, \ldots n)$ and $v_i$
Adjust pulse frequency $f_i$ at $x_i$
Initialize pulse rates $r_i$ and the loudness $A_i$
While the maximum number of iterations is not fulfilled:
Produce a new solution by tuning frequency,
and updating velocities and locations /solutions
$f_i = f_{min} + (f_{max} - f_{min})\beta$
$v_i^{t+1} = v_i^t + (x_i^t - x_*)f_i$
$x_i^{t+1} = x_i^t + v_i^{t+1}$
   If (rand (0, 1) > $r_i$ )
      Select a solution among the best solutions
      Generate a local solution around the selected best solution (Equation 4)
   End if
   Generate a new solution by flying randomly
   If (rand (0, 1) < $A_i$ and $f(x_i) < f(x_*)$)
      Accept the new solutions
      Increase $r_i$ and decrease $A_i$ (Equations 5 and 6)
   End if
The new population evaluated using the objective function $f(x)$
Rank the bats and find the current best $x_*$
End while

In the local search phase, when a solution is selected among the current best solutions, a random walk is used to generate a new solution for each bat.

$$x_{new} = x_{old} + \varepsilon A^t \qquad (4)$$

Where $\varepsilon$ is a random number between -1 and 1, while $A^t = (A_i^t)$ refers to the average value of loudness at ($t$) iteration. Moreover, when the iterations progressed, the loudness ($A_i$) and the rate ($r_i$) of pulse emission should be changed according to Equations (5) and (6).

$$A_i^{t+1} = \alpha A_i^t \qquad (5)$$
$$r_i^{t+1} = r_i^0[1 - \exp(-\gamma t)] \qquad (6)$$

When the bats find their prey, the value of loudness decreases, whereas the rate of pulse emission increases. These updates only occur when the new solutions are enhanced, which means that they are changing their positions toward the optimal solution [26].

## 2.2. Key Features of Bat Algorithm

The balance between the exploration phase (diversification) and the exploitation phase (intensification) of the search mechanism is regarded as one of the most significant aspects of a population-based metaheuristic. A global search is done by exploration. At the same time, exploitation is responsible for local search. In BA, like other population-based algorithms, several parameters drive their search behaviors, such as population number, the maximum number of iterations, and problem dimension, in the initial step of the bat algorithm, a random value of (A) and (r) is assigned to each bat. Usually, a value of A ranges between [2, 0] while the value of r ranges between [1, 0]. With the progress of the algorithm, as the bats get closer to the optimal solution, the value of (A) decreases and becomes 0 at the ideal solution, while the value of (r) increases to become 1 at the ideal solution. Figures 1 and 2 illustrate this situation. So, from the standard pseudo-code of the bat algorithm, we conclude that this algorithm is robust at the exploitation phase and bad at exploration because, as seen in Figure 2, the increase rate of (r) is proportional to the number of iterations in the beginning, iterations of the algorithm, the value of pulse rates ($r_i$) is tuned to zero, and the possibility of selecting a random number range [0, 1] greater than ($r_i$) is highly ensured by a bat, thus, the algorithm loses the exploration ability and easily get trapped into some local optima, therefore it cannot search globally well and as the algorithm performs the search depending on random walks, so a rapid convergence cannot be guaranteed. The BA also does not store the position of the best solution found so far during the optimization process, which causes bats to sometimes tend to move away from the promising area of solutions search space. This study aims to modify the standard Bat Algorithm to eliminate the above-mentioned drawbacks [28].

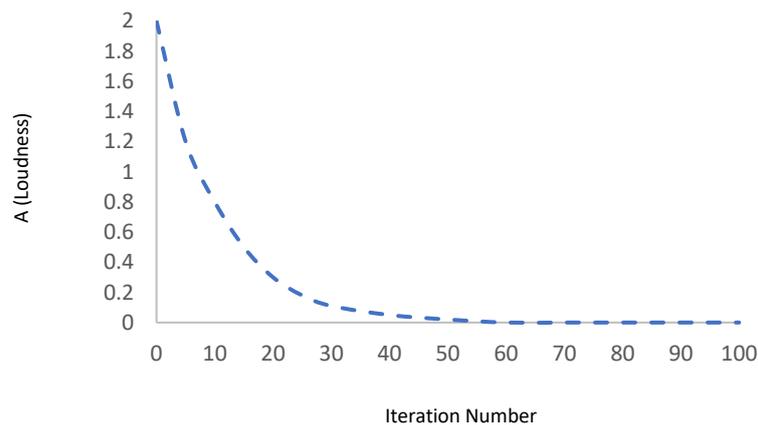

**Figure 2: The value of loudness (A) during 100 iterations**

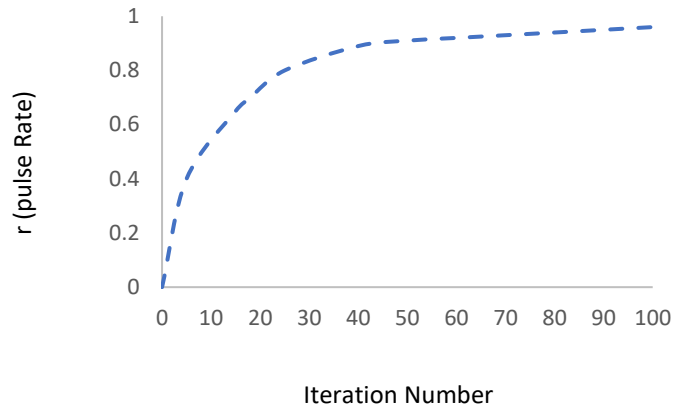

Figure 3: The value of pulse rate (r) during 100 iterations

## 3. The Proposed Bat Algorithm

The original Bat algorithm suffers from the diversity issue. That is the agents are finding the areas in the search space; however, they do not have the chance to aggregate around those areas, and hence the algorithm fails to converge during the iterations [29], [30].

Therefore, there should be an approach to force the agents to move towards the global best. Therefore, our suggested modifications to the Bat Algorithm have included two parts, which are the bat's new position and the velocity equations:

1) In the standard BA, to find the velocity of the new solution, the velocity, and the frequency of the current solution are used as presented in Equation (2). In the MBA, to direct the search toward the optimal solution, the velocity and the frequency of the best solution found so far are also used (Equation 7).

2) In the standard BA, to find the position of the new solution, only the position of the current solution and the velocity of the new solution are used as shown in Equation (3). In the MBA, rather than using the position of the current solution, the position of the best solution is used and the velocity of the current solution is multiplied by the frequency of the best solution found so far (Equation 8). The pseudo-code of the modified BA is listed below in Algorithm (2).

**Algorithm (2): The Pseudo-code of the Modified BAT algorithm**

Objective Function $f(x)$, $x = (x_1, \ldots x_d)^T$
Initialize the bat individuals $x_i$, $(i = 1, 2, \ldots n)$ and $v_i$
Set pulse frequency $f_i$ at $x_i$
Initialize pulse rates $r_i$ and the loudness $A_i$
**While not** the maximum number of iterations is satisfied:
Generate new solution by adjusting frequency,
and updating velocities and locations /solutions
$f_i = f_{min} + (f_{max} - f_{min})\beta$
$v_i^{t+1} = v_i^t . f_i^t - (v_* . f_*) + (x_i^t - x_*).f_*$
$x_i^{t+1} = x_* + (v_i^{t+1}.f_*)$
**If** (rand(0, 1) > $r_i$ )
   Select a solution among the best solutions
   Generate a local solution around the selected best solution (Equation 4)
**End if**
Generate a new solution by flying randomly

**If** (rand (0, 1) < $A_i$ and $f(x_i) < f(x_*)$)
   Accept the new solutions
   Save the frequency and the velocity of the best new solution as $f_*$ and $v_*$
   Increase $r_i$ and decrease $A_i$ (Equations 5 and 6)
**End if**
The new population is evaluated using the objective function $f(x)$
Rank the bats and find the current best $x_*$
**End while**

In the proposed algorithm, the velocity and the new position equations have been modified as shown in Equations (7) and (8) to eliminate the explained problem, also in the algorithm after evaluating the objective function of the new positions, the value of the frequency and the velocity of the best solution are saved (If there was an improvement).

$$v_i^{t+1} = v_i^t . f_i^t - (v_* . f_*) + (x_i^t - x_*). f_* \qquad (7)$$

$$x_i^{t+1} = x_* + (v_i^{t+1}. f_*) \qquad (8)$$

## 4. Materials and Methods

In this section, the benchmark testing functions, parameter setting, and evaluation criteria of the experiment are presented.

### 4.1. Benchmark Testing Functions

Benchmark test functions have various characteristics; unimodal benchmark functions and multimodal benchmark functions. Unimodal benchmark functions have a single optimum and are used to evaluate the exploitation phase and the rate of algorithm convergence. Multimodal functions have several optimal solutions and are used to test the exploration phase of the algorithm. First, 23 mathematical benchmark functions have been performed to test the performance of the proposed algorithm [27]. These functions are categorized into two sets: unimodal benchmark functions (*f1-f7*) and multimodal benchmark functions (*f8-f23*). Secondly, a set of 25 CEC2005 benchmark functions has been used as an additional evaluation of the algorithm [31]. CEC benchmarks are proposed within the IEEE Congress on Evolutionary Computation. CEC2005 includes four types of benchmark functions; these are unimodal functions, multimodal functions, expanded multimodal functions, and hybrid composition functions. Finally, a group of 10 CEC2019 benchmark functions has been implemented [32].

### 4.2. Parameter Setting

To obtain more feasible results for a given problem, it is necessary to use proper values to the selected parameters, also to get fair and accurate results for all compared algorithms, the same parameter tuning is used and the initial population is selected randomly, for instance, the population size is 30 bats, the algorithm is allowed to find the best optimum solution in 500 generations, while the dimension is 30. To find statistical measures like the average and the standard deviation, the algorithm was executed 30 times.

### 4.3. Evaluation Criteria

To get a better comparison, three ways are used to evaluate algorithms, these standard evaluations are:
1) Calculating the average and standard deviation of the optimum solutions.
2) Comparing the standard Bat algorithm with the MBA algorithm by building a box and whisker plot.

Comparing the MBA with other metaheuristic algorithms.

### 5. Result and Evaluation

This section presents the evaluation results of the MBA on different types of benchmark testing functions. Meanwhile, the exploitation and exploration of MBA is measured. Lastly, the complexity of the MBA is computed.

### 5.1. Results

In the classical benchmark functions, there are two groups of functions, the first group is unimodal functions, which have only one optimum global solution and are used to validate the exploitation ability and the convergence proportion of the proposed algorithm. While, multimodal benchmark functions have more than one optimal solution, and they are used to check the exploration level and also to avoid trapping into the local optima. By evaluating these functions, it could be easy to test the exploitation and exploration ability of the modified algorithm.

Table (1) presents the results of MBA are better than the standard bat algorithm for the majority of the functions (the signs "+", "-", and "*" at the rightmost column named "significant" in all tables denotes that MBA is better than BA, MBA is worse than BA, and MBA is equal to BA respectively).

Table (1): Evaluation results of standard BA and MBA on Classical Benchmark Functions

| Function | BA | | MBA | | Significant |
|---|---|---|---|---|---|
| | Average | STD | Average | STD | |
| F1 | 1.622E+01 | 1.206E+01 | **6.433E-05** | **3.217E-04** | + |
| F2 | 1.745E+01 | 8.761E+00 | **3.022E+00** | **3.092E+00** | + |
| F3 | 4.032E+02 | 6.548E+02 | **0.000E+00** | **0.000E+00** | + |
| F4 | 8.270E-01 | 2.004E-01 | **0.000E+00** | **0.000E+00** | + |
| F5 | 2.222E+03 | 1.211E+03 | **2.760E+01** | **5.543E+00** | + |
| F6 | 1.419E+01 | 1.339E+01 | **2.023E+00** | **1.848E+00** | + |
| F7 | 4.803E+01 | 9.176E+01 | **1.316E-04** | **1.012E-04** | + |
| F8 | -5.288E+01 | 1.138E+01 | **-1.171E+02** | **1.227E+00** | + |
| F9 | 2.017E+01 | 1.293E+01 | **0.000E+00** | **0.000E+00** | + |
| F10 | 2.378E+00 | 1.421E+00 | **8.882E-16** | **0.000E+00** | + |
| F11 | 6.352E-01 | 2.567E-01 | **0.000E+00** | **0.000E+00** | + |
| F12 | 2.510E+00 | 1.617E+00 | **1.425E-01** | **1.612E-01** | + |
| F13 | 1.041E+00 | **9.840E-02** | 7.165E-01 | 4.262E-01 | + |
| F14 | 1.268E+01 | 2.650E-02 | **1.267E+01** | **2.392E-10** | + |
| F15 | 1.340E-02 | 1.470E-02 | **6.500E-03** | **4.900E-03** | + |
| F16 | -6.361E-01 | **2.542E-01** | -6.985E-01 | 2.884E-01 | + |
| F17 | 8.942E+00 | 2.482E+00 | **6.369E-01** | **1.700E-01** | + |
| F18 | 2.423E+02 | 2.048E+02 | **8.464E+01** | **3.599E+01** | + |

| F19 | -2.322E+00 | **7.442E-01** | -2.731E+00 | 8.442E-01 | + |
| F20 | -3.658E-01 | **4.012E-01** | -6.907E-01 | 5.140E-01 | + |
| F21 | -2.804E+00 | 1.476E+00 | -3.006E+00 | 1.168E+00 | + |
| F22 | -2.997E+00 | 1.042E+00 | -3.331E+00 | 1.000E+00 | + |
| F23 | **-3.251E+00** | 1.813E+00 | -3.124E+00 | 9.905E-01 | - |

Another evaluation has been done using 25 benchmark functions of CEC2005. Table (2) depicts the comparison results of the standard Bat algorithm and the proposed algorithm. MBA outperformed the original algorithm in *F1, F2, F3, F4, F5, F7, F9, F10, F12, F13, F16, F17, F18, F22, F23, F24*, and F25. While BA has better performance in the remaining functions (*F11, F15, F19, F20, and F21*), both two algorithms have the same performance in three functions (*F6, F8, and F14*). As a result, it can be said that the modified algorithm enhanced the performance of the original Bat algorithm in about 17 functions.

Table (2): Evaluation results of MBA against the standard BA on CEC2005

| Function | BA | | MBA | | Significant |
|---|---|---|---|---|---|
| | Average | STD | Average | STD | |
| F1 | 8.92E+04 | 153.2354 | **8.91E+04** | 140.473 | + |
| F2 | 1.19E+06 | 5.19E+04 | **8.28E+04** | 5.27E+02 | + |
| F3 | 3.08E+09 | 3.12E+07 | **2.30E+09** | 1.27E+07 | + |
| F4 | 1.28E+06 | **5.20E+04** | 7.00E+05 | 8.63E+04 | + |
| F5 | 6.88E+04 | **93.867** | 6.67E+04 | 344.4199 | + |
| F6 | 4.35E+10 | 1.25E+08 | 4.35E+10 | 1.66E+08 | * |
| F7 | 4.86E+03 | 8.6873 | **3.76E+03** | 80.7196 | + |
| F8 | 21.2409 | 0.0633 | 21.2838 | 0.0925 | * |
| F9 | 506.0128 | 22.5011 | **486.9987** | 24.0675 | + |
| F10 | 918.9271 | 27.9841 | **842.593** | 49.9419 | + |
| F11 | **46.7074** | 2.1821 | 47.5323 | 1.9614 | - |
| F12 | 2.27E+06 | 2.29E+05 | **2.03E+06** | 2.98E+05 | + |
| F13 | 4.32E+03 | 3.25E+03 | **83.1821** | 51.117 | + |
| F14 | 14.7553 | 0.0348 | 14.7597 | 0.0391 | * |
| F15 | **824.2841** | 151.0182 | 832.848 | 104.7229 | - |
| F16 | 629.5401 | 127.3081 | **582.1609** | 123.7847 | + |
| F17 | 635.9273 | 103.2046 | **587.2459** | 143.6904 | + |
| F18 | 354.0225 | 170.4205 | **324.7698** | 164.7771 | + |
| F19 | **329.3632** | 187.2904 | 374.0069 | 183.458 | - |
| F20 | **309.691** | 158.7835 | 347.4832 | 188.2756 | - |
| F21 | **1.11E+03** | 270.5426 | 1.13E+03 | 259.37 | - |
| F22 | 1.14E+03 | 400.9433 | **867.0604** | 435.6023 | + |
| F23 | 1.17E+03 | 294.569 | **784.1982** | 531.3192 | + |
| F24 | 825.0976 | 222.5496 | **818.8262** | 123.2495 | + |
| F25 | 879.9396 | 220.2351 | **860.2515** | 130.3779 | + |

Finally, the CEC2019 was used to test the performance of the MBA and BA algorithms. Table (3) presents that the MBA has lower average results compared to BA in 7 out of 10 functions. However, MBA is not very competitive with BA in just two functions (*F6 and F10*), while both algorithms have similar results in just one function *(F3)*. Overall, we can say that MBSA would outperform the standard Bat algorithm in most CEC2019 benchmark functions.

Table (3) Comparison Results of standard BA and MBA on CEC2019

| Function | BA | | MBA | | Significant |
|---|---|---|---|---|---|
| | Average | STD | Average | STD | |
| F1 | 8.04E+05 | 2.88E+05 | **4.26E+05** | 2.31E+05 | + |
| F2 | 2.02E+01 | 6.18E-01 | **1.85E+01** | 1.47E-01 | + |
| F3 | 1.27E+01 | 1.30E-03 | 1.27E+01 | 1.20E-03 | * |
| F4 | 7.14E+04 | 3.55E+02 | **5.39E+04** | 1.57E+03 | + |
| F5 | 8.57E+00 | 2.23E-02 | **8.14E+00** | 2.45E-01 | + |
| F6 | **1.42E+01** | 8.00E-01 | 1.45E+01 | 1.42E+00 | - |
| F7 | 4.60E+03 | 1.29E+02 | **4.52E+03** | 5.36E+02 | + |
| F8 | 8.39E+00 | 3.83E-01 | **8.35E+00** | 5.51E-01 | + |
| F9 | 6.44E+03 | 6.62E+01 | **4.67E+03** | 7.02E+01 | + |
| F10 | **2.20E+01** | 1.74E-01 | 2.21E+01 | 1.60E-01 | - |

## 5.2. Evaluations

From the original BA, it may be mentioned here that the MBA was already shown reasonable to the other advanced optimization algorithms (PSO and GA). Hence, the proposed MBA has also exhibited competitive results against the mentioned algorithms as shown in Table (4). These algorithms are tested on a set of well-known benchmark functions, which include De Jong's standard shifted sphere function, Schwefel problem, Rosebrock's function, Generalized Rastrigin's, Ackley's function, and Generalized Griewangk's. For more accurate results, the algorithms in Table (4) have been executed 30 times with the same population size (40); every time, the algorithm searches the optimal solution in the search space in 500 iterations, then the statistical measurements (standard deviation and the average) are calculated. The results indicate that PSO outperforms genetic algorithms, while the MBA Algorithm is much superior to both algorithms in terms of accuracy and efficiency.

Table (4): Comparison Results of MBA with GA and PSO

| Function Name | GA | | PSO | | MBA | |
|---|---|---|---|---|---|---|
| | Average | STD | Average | STD | Average | STD |
| De Jong's shifted sphere (d=256) | 2.541E+04 | 1.24E+03 | 1.704E+04 | 1.123E+03 | **3.017E+02** | 11.3423 |
| Schwefel problem (d=128) | 2.273E+05 | 7.572E+03 | 1.452E+04 | 1.275E+03 | **0.000E+00** | 0.00E+00 |
| Generalized Rosenbrock's (d=16) | 5.572E+04 | 8.901E+03 | 3.276E+04 | 5.325E+03 | **1.370E+01** | 3.72E+00 |
| Generalized Rastrigin's | 1.105E+05 | 5.199E+03 | 7.949E+04 | 3.715E+03 | **1.0144** | 10.1439 |
| Ackley's function (d=128) | 3.272E+04 | 3.327E+03 | 2.341E+04 | 4.325E+03 | **2.523E-04** | 2.50E-03 |
| Generalized Griewangk's | 7.093E+04 | 7.652E+03 | 5.597E+04 | 4.223E+03 | **0.000E+00** | 0.00E+00 |

MBA also is compared with one of the modern optimizations algorithms, which is DA, the reason for selecting this algorithm due to two points: First, it is proven to have an outstanding performance both on benchmark test functions and also to solve real-world problems, the second reason is that this algorithm has proven its superiority over the PSO algorithm and GA, and thus, we have proved the superiority of our proposed algorithm (MBA) over DA algorithm, so, this means its superiority over the mentioned algorithms too. To get fair results, both competitors use the same parameter setting as the settings used in their original papers [2]. Also, to collect quantitative outputs and to find the average and the standard deviation of the best optimum solutions, both algorithms are run on the standard benchmark functions 25 times, for 500 iterations using 40 search agents. As shown in Table (5), per the results of applying the MBA and DA on the unimodal test functions (*F1–F7*), it is clear that the MBA algorithm outperforms DA on all these functions, while applying the algorithms on multi-modal test functions (*F8-F13*) displays that again the MBA algorithm shows significantly better results than DA in (*F9, F10, ,F11, F12,* and *F13*). However, DA is only outperformed in (*F8*). The results of composite test functions (*F14-F23*) show that both algorithms have about the same performance in *(F15, F16, F17*, and *F19*). However, the MBA shows that superiority is not as significant as those of unimodal and multimodal test functions, this is due to the complexity of these functions.

Table (5): Comparison Results of MBA with DA

| Function | MBA | | DA | | Significant |
|---|---|---|---|---|---|
| | Average | STD | Average | STD | |
| F1 | **6.43E-05** | 3.22E-04 | 1.89E+00 | 3.87E+00 | + |
| F2 | **8.15E-01** | 9.18E-01 | 1.03E+00 | 9.50E-01 | + |
| F3 | **0.00E+00** | 0.00E+00 | 6.00E+01 | 8.56E+01 | + |
| F4 | **0.00E+00** | 0.00E+00 | 2.08E+00 | 1.99E+00 | + |
| F5 | **7.23E+00** | 3.44E+00 | 1.15E+04 | 3.00E+04 | + |
| F6 | **8.14E-01** | 5.30E-01 | 3.82E+00 | 6.46E+00 | + |
| F7 | **9.34E-04** | 3.10E-03 | 3.32E-02 | 3.90E-02 | + |
| F8 | -3.91E+01 | 3.23E-01 | **-2.89E+03** | 4.21E+02 | - |
| F9 | **0.00E+00** | 0.00E+00 | 2.92E+01 | 1.34E+01 | + |
| F10 | **8.88E-16** | 0.00E+00 | 2.38E+00 | 1.13E+00 | + |
| F11 | **0.00E+00** | 0.00E+00 | 5.46E-01 | 4.07E-01 | + |
| F12 | **1.52E-01** | 3.78E-01 | 1.27E+00 | 1.18E+00 | + |
| F13 | **1.80E-01** | 8.00E-02 | 9.17E-01 | 2.05E+00 | + |
| F14 | 1.27E+01 | 2.11E-10 | **1.36E+00** | 6.33E-01 | - |
| F15 | 6.20E-03 | 5.80E-03 | 4.40E-03 | 6.40E-03 | * |
| F16 | -7.09E-01 | 2.89E-01 | -1.03E+00 | 3.00E-09 | * |
| F17 | 6.62E-01 | 2.07E-01 | 3.98E-01 | 1.79E-07 | * |
| F18 | 7.64E+01 | 2.65E+01 | **3.00E+00** | 3.99E-15 | - |
| F19 | -3.01E+00 | 4.86E-01 | -3.86E+00 | 1.41E-04 | * |
| F20 | -9.38E-01 | 6.48E-01 | **-3.24E+00** | 9.26E-02 | - |
| F21 | -3.04E+00 | 9.92E-01 | **-7.83E+00** | 2.72E+00 | - |
| F22 | -3.11E+00 | 1.01E+00 | **-8.65E+00** | 2.61E+00 | - |
| F23 | -2.89E+00 | 8.29E-01 | **-7.84E+00** | 2.91E+00 | - |

Also, to support the significance of the results presented in Table (5), the *p* values of the Wilcoxon rank-sum test are calculated and the results of a statistical comparison are shown in Table (6). (Wilcoxon rank-sum tests are used to determine whether two samples are likely to have come from the same two underlying populations that have the same mean).

Table (6) THE WILCOXON RANK-SUM TEST FOR CLASSICAL BENCHMARKS

| Function | P values of MBA vs. DA |
|---|---|
| F1 | **2.226E-02** |
| F2 | 5.039E-01 |
| F3 | **1.823E-03** |
| F4 | **2.232E-05** |
| F5 | 6.746E-02 |
| F6 | **2.890E-02** |
| F7 | **4.166E-04** |
| F8 | **9.331E-22** |
| F9 | **9.202E-11** |
| F10 | **1.897E-10** |
| F11 | **6.138E-07** |
| F12 | **1.509E-04** |
| F13 | 8.287E-02 |
| F14 | **8.438E-32** |
| F15 | **3.264E-02** |
| F16 | **9.917E-06** |
| F17 | **1.312E-06** |
| F18 | **6.309E-13** |
| F19 | **5.923E-09** |
| F20 | **3.612E-15** |
| F21 | **1.723E-09** |
| F22 | **8.125E-10** |
| F23 | **1.534E-08** |

In order to obtain an additional evaluation of the suggested algorithm, it was compared with another modified and enhanced bat algorithm. First, in Fractional Lévy Flight Bat Algorithm (FLFBA) [33] where Fractional Levy Flights (FLF) was combined with the DE algorithm to enhance the performance of the standard Bat algorithm, the experimental results indicates the outperforming of MBA in the majority benchmark functions, as shown in Table (7), also Chimp Optimization Algorithm (ChOA) [34] has been compared with MBA, it is noticeable from Table (7) that MBA is better in 12 functions among 23.

Table (7) Comparison Results of MBA with FLFBA and ChOA

| Function | MBA | | FLFBA | | ChOA | |
|---|---|---|---|---|---|---|
| | Average | STD | Average | STD | Average | STD |
| F1 | 6.43E-05 | 3.22E-04 | 3.29E+05 | 1.61E+05 | 2.20E-18 | 7.24E-18 |
| F2 | 8.15E-01 | 9.18E-01 | 3.09E+03 | 2.04E+03 | 1.56E-12 | 3.35E-12 |
| F3 | 0.00E+00 | 0.00E+00 | 3.90E+05 | 3.18E+05 | 8.17E-07 | 3.72E-06 |
| F4 | 0.00E+00 | 0.00E+00 | 3.37E+05 | 2.27E+05 | 4.93E-06 | 1.11E-05 |
| F5 | 7.23E+00 | 3.44E+00 | 2.61E+04 | 1.35E+04 | 8.93E+00 | 1.72E-01 |
| F6 | 8.14E-01 | 5.30E-01 | 3.54E+05 | 2.33E+05 | 2.18E-01 | 2.03E-01 |
| F7 | 1.32E-04 | 3.10E-03 | 4.61E+01 | 4.39E+01 | 8.13E-04 | 8.07E-04 |
| F8 | -1.17E+02 | 3.23E-01 | 7.79E+06 | 5.03E+06 | -2.21E+03 | 7.78E+01 |
| F9 | 0.00E+00 | 0.00E+00 | 8.61E+02 | 5.80E+02 | 3.66E+00 | 4.51E+00 |
| F10 | 8.88E-16 | 0.00E+00 | 3.96E+04 | 2.42E+04 | 1.93E+01 | 2.64E+00 |
| F11 | 0.00E+00 | 0.00E+00 | 1.26E+07 | 8.88E+06 | 7.01E-02 | 7.88E-02 |
| F12 | 1.43E-01 | 3.78E-01 | 6.86E+04 | 3.87E+04 | 3.78E-02 | 1.53E-02 |
| F13 | 1.80E-01 | 8.00E-02 | 6.80E+04 | 3.71E+04 | 9.35E-01 | 9.30E-02 |
| F14 | 1.27E+01 | 2.11E-10 | 2.85E-09 | 7.79E-09 | 1.32E+00 | 1.78E+00 |
| F15 | 4.20E-04 | 5.80E-03 | 4.65E-04 | 5.42E-04 | 1.32E-03 | 5.46E-05 |
| F16 | -7.09E-01 | 2.89E-01 | 1.10E-11 | 2.91E-11 | -1.03E+00 | 1.32E-05 |
| F17 | 6.62E-01 | 2.07E-01 | 3.21E-11 | 7.92E-11 | 3.04E-01 | 2.33E-06 |
| F18 | 7.64E+01 | 2.65E+01 | 1.13E-12 | 1.79E-12 | 3.00E+00 | 2.11E-04 |
| F19 | -3.01E+00 | 4.86E-01 | 0.00E+00 | 0.00E+00 | -3.85E+00 | 1.79E-03 |
| F20 | -9.38E-01 | 6.48E-01 | 0.00E+00 | 0.00E+00 | -2.57E+00 | 5.68E-01 |
| F21 | -3.04E+00 | 9.92E-01 | 0.00E+00 | 0.00E+00 | -3.48E+00 | 2.01E+00 |
| F22 | -9.33E+00 | 1.01E+00 | 0.00E+00 | 0.00E+00 | -3.85E+00 | 2.04E+00 |
| F23 | -5.12E+00 | 8.29E-01 | 0.00E+00 | 0.00E+00 | -4.24E+00 | 2.03E+00 |

## 5.3. Exploration and Exploitation Measurement

Exploration and exploitation are the two primary search behaviors that swarm members often engage in. In the first, the individuals are separating from one another and the spaces between them are growing wider. It is at this phase that fresh places are found and potential local optima traps are avoided. The individuals, on the other hand, are intensifying and getting closer together throughout the exploitation period. During this stage, individuals prefer to look locally in their immediate area and converge on the global optimum. Preventing early convergence in metaheuristic algorithms requires achieving the right balance between the exploration and exploitation stages [35].

The dimension-wise diversity measurement is used in this work to quantitatively assess the level of exploitation and exploration of the chosen algorithms. Additionally, this study used the median instead of the mean in Equation (9) since it more properly captures the population's center [36].

$$Div_j = \frac{1}{n}\sum_{i=1}^{n} meadian\,(x^j) - x_i^j;  \qquad (9)$$

$$Div = \frac{1}{D}\sum_{j=1}^{D} Div_j  \qquad (10)$$

Where median $(x^j)$ is the median of dimension j in the whole population. $x_i^j$ represents the dimension $j$ of the individual $i$. $n$ depicts the population size and D depicts the dimension.

The diversity in each dimension $Div_j$ can be expressed as the average distance between each search agent's dimension j and that dimension's median. Then, In Div, the average diversity across all dimensions is calculated. By averaging the equation (10), it is possible to determine an algorithm's exploration and exploitation percentage:

$$XPL\% = \left(\frac{Div}{Div_{max}}\right) * 100  \qquad (10)$$

$$XPT\% = \left(\frac{|Div - Div_{max}|}{Div_{max}}\right) * 100  \qquad (11)$$

Where $Div_{max}$ represent the maximum diversity, value achieved in the entire optimization process. *XPL%* and *XPT %* are the degree of exploration and exploitation respectively.

Table (8) The ratio of exploration and exploitation between the Bat Algorithm and the proposed MBA on CEC2019

| Function | BA | | MBA | |
|---|---|---|---|---|
| | Exploration% | Exploitation% | Exploration% | Exploitation% |
| 1 | 53.0484 | 46.9516 | 30.3407 | 69.6593 |
| 2 | 36.0768 | 63.9232 | 62.9744 | 37.0256 |
| 3 | 39.2078 | 60.7922 | 51.1434 | 48.8566 |
| 4 | 58.5991 | 41.4009 | 4.2517 | 95.7483 |
| 5 | 48.7755 | 51.2245 | 13.2991 | 86.7009 |
| 6 | 65.4521 | 34.5479 | 14.1305 | 85.8695 |
| 7 | 59.6302 | 40.3698 | 6.6598 | 93.3402 |
| 8 | 14.3565 | 85.6435 | 9.0318 | 90.9682 |
| 9 | 48.6386 | 51.3614 | 57.5594 | 42.4406 |
| 10 | 70.595 | 29.405 | 10.9647 | 89.0353 |

As it can be seen from Table (8), the proposed algorithm has been successful in solving the diversity issue and the exploitation ratio has increased for most of the test functions. Having said that, it is worth mentioning, this is not entirely true for all test functions and more investigation regarding the structure and operation of the algorithm is required. Figure 4 Shows the exploration and exploitation ratio for some of the test functions throughout the iterations.

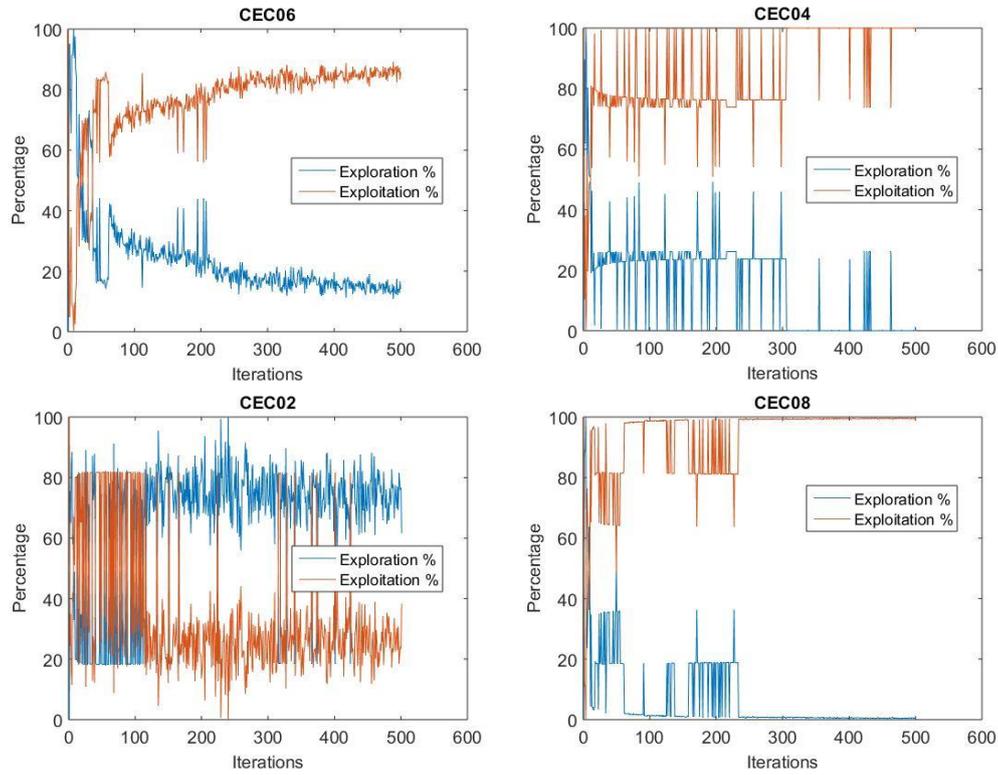

Figure 4: The exploration and exploitation ratio of the MBA for some test functions

**5.4. Computational Complexity**

Regarding the computational complexity of MBA, Since the proposed modifications are linear in nature and do not need additional computation or the evaluation of a fitness function, they do not change the original Bat algorithm's computation complexity in the context of big O notation.

therefore, the complexity of the proposed algorithm for each iteration is $O(ND + N * fit)$, where $N$ is the population size, $D$ is the dimension of the problem's search space, and $fit$ is the fitness value of the objective function. However, the execution time of the proposed algorithm has slightly increased as shown in Table (9), where the original Bat algorithm is faster by 0.2 second and this can be seen as a limitation of the proposed algorithm. The simulation experiments have been conducted on an 11th Gen Processor Intel(R) Core (TM) i7-1195G7 @ 2.90GHz, (32G RAM).

Table (9): Comparison of execution time between BA and MBA algorithms on CEC2019

| Function | Execution time | |
|---|---|---|
| | BA | MBA |
| 1 | 4.0008 | 4.368042 |
| 2 | 0.909373 | 1.363406 |
| 3 | 1.054642 | 1.25811 |
| 4 | 1.309201 | 1.368091 |
| 5 | 1.023952 | 1.593415 |
| 6 | 2.562457 | 2.606514 |
| 7 | 2.758072 | 2.813905 |
| 8 | 1.39392 | 1.697544 |
| 9 | 1.215297 | 1.458978 |
| 10 | 1.222529 | 1.24553 |

## 6. MBA for Business Process Optimization

MBA can be used to solve one of the real-time critical problems, which is the inbound call center problem. This section is about a case study of MBA for call center, and utilizing MBA on real-world application.

### 6.1. A Case Study of MBA for Call Center

The agents in the call center receive about (200,000) calls in a day and they have (2700) agents working 8 hours over the day in 3 different shifts. For the call center, call handle time is one of the most significant business factors, which consists of hold time (wait time), call time, and post-call work time. The specific call center has about 12 minutes for call handle time, which includes the wait time of the customer call close to 4 minutes and 8 minutes for the call time itself. The average (daily) available time for agents is (1,296,000) minutes (equivalent to 21,600 hours in a day), while the required call time is about (1, 600, 00 minutes) (equivalent to 26,666 hours in a day), so the difference between the available and the required time is huge. To resource management, time handling is regarded as an important business factor. The call handle time is presented in Equation (12).

$$Call\ handle\ time\ = \frac{Hold\ time\ +\ Call\ time\ +\ Post\ call\ time}{total\ calls} \quad (12)$$

Whereas, the hold time is the time that the caller waits before being linked to the respective agent, the call time is the time required between the caller and the agent and the post-call time is the time for termination of the call-related activities after the call.

MBA optimizer aims to assign the callers to the right agents so that the overall call time can be minimized from 8 to about 5.5 to create enough time for agents to take coffee breaks. The first step in the solution is to assign the right agents with suitable callers based on their profiles and the tendency to have a compatible conversation. Figure 5 below explains how this intelligent (Caller-Agent connector) system works.

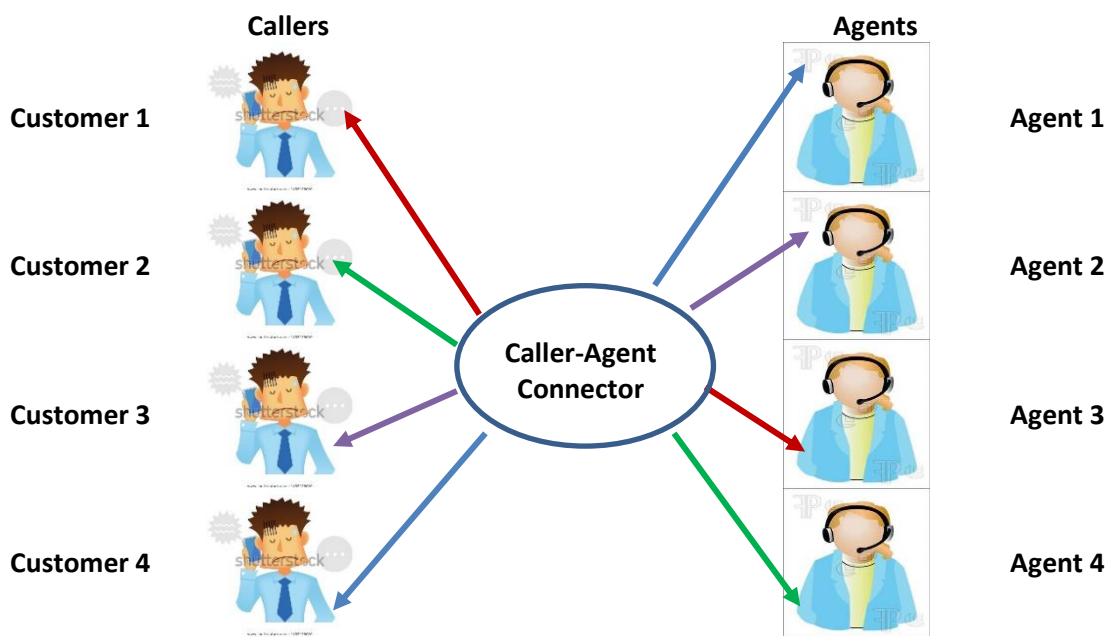

Figure 5: Automated Intelligence to connect callers with agents

## 6.2. MBA for Real-world Application

Optimization can be defined as a method to achieve some objectives optimally or to optimize something, such as cost, time, quality, performance, or productivity. In real-world applications, these goals are always limited, so the new metaheuristic algorithms were used to optimally use these resources under various constraints, such as GA, ACO, PSO, BA or DE.

Any optimization problem consists of three basic components:

 a) Objective/Goal.
 b) Decision Variables.
 c) Constraints (generally resource constraints).

Like any other Meta-heuristic algorithm, an MBA can be used to solve a real-world task assignment problem (call center problem). In business process optimization, the aim is how to assign some workers on a one-to-one basis so that the jobs are completed in the least time or at the least cost. Our optimization problem case study example is a call center for a digital television network, we optimized one of the key metrics of this problem, which is called handle time to achieve an intelligent real-world optimizer that will connect each caller to the right agent based on minimum query resolution time. This will be done with the following steps:

 a) Unsupervised learning: Each caller has its demographics, usage, and complaints of log information. In the first step of the solution, unsupervised learning is applied to the data set to divide the callers into different profiles based on their information.
 b) Supervised learning: - supervised learning methods are used to determine an agent's propensity to resolve a query for different caller profiles within the required query time.
 c) Meta-heuristic (Our Modified Bat Algorithm) analytics is performed to link the right caller profile segmented in step 1 with the right agent profile derives in step 2.

To explain our optimizer, we used a simple case of connecting many callers with the same number of available agents as shown in Figure 5, from the first two steps, we obtained the average time (in seconds) that each agent takes to answer and manage the queries for the caller profiles as shown in Table (9). The maximum query resolution time for the table is 2343 (about 9 minutes and 45 seconds) average call time. If we have $n$ callers with $n$ agents, there will be ($n*n$) possibilities of connections, and $(n!)$ combinations of the callers-agent's connections, for example, if we have 4 callers with 4 agents, there will be 16 possibilities of connections with 24 combinations of the callers-agent's connections (like C1A1 + C2A2+ C3A3+ C4A4). Thus, if we have 5, 6, 7, and 8 callers (at the same time 5, 6, 7, and 8 agents), there will be 120, 720, 5040, and 40320 combinations of possible connections respectively. So, when the number of callers increases, the probability of the search space for the optimum solution (minimum call time) also increases, therefore it will be computationally incredible to search all these probabilities using traditional linear programming methods and it will consume infinitely duration of computations. Therefore, our MBA is used to optimize this problem, by approaching the average call time to the optimal solution (246+436+324+157=1163 seconds), which equals 4 minutes and 51 seconds. This result shows a significant decrease from 8 minutes of average call time.

Table (9): Callers-Agents average call time in seconds

|  | Agent 1 | Agent 2 | Agent 3 | Agent 4 |
|---|---|---|---|---|
| Caller Profile 1 | 540 | 215 | 221 | 246 |
| Caller Profile 2 | 848 | 436 | 542 | 936 |
| Caller Profile 3 | 324 | 81 | 288 | 328 |
| Caller profile 4 | 775 | 579 | 157 | 263 |

Additionally, MBA is applied to this problem with 30 search agents for 1000 generations, and the results of average call time can be seen in Figure 6. At this point, MBA is better than the original bat algorithm, with a difference of about one minute during a single connection, and thus leads to save several minutes during the working day, which agents can benefit from taking a break and drinking coffee.

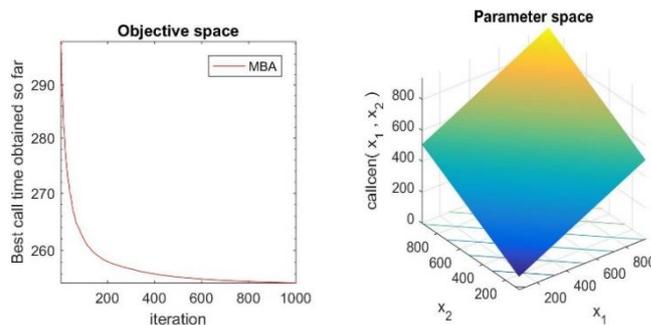

Figure 6: The average call time results for 1000 iterations with 30 bat search agents on the Call Center problem

## 7. Conclusion

In this study, the standard BA was explained briefly, and characteristics, functionality, and echolocation capability were presented. BA has restricted capacity for exploration and is prone to being influenced by local optima. On that basis, this work suggested an algorithm, called MBA, as a solution to overcome the limitation of local optima found in the original BA. A group of 23 single objective benchmark functions (Unimodal, multimodal, and composite test functions) was used to evaluate the performance of the MBA. Also, MBA tested on another 25 functions, which are called CEC2005 benchmark functions. And finally, a set of 10 modern CEC2019 benchmark functions was also implemented. The MBA was also compared with the most popular meta-heuristic algorithm (PSO and GA). MBA outperformed the standard algorithms in most cases of benchmark functions and produced comparative results on the others. Also, when comparing MBA with other optimization algorithms, such as PSO and GA, it showed that it was much superior to both algorithms in terms of accuracy and efficiency. Also, MBA compared with the DA as one of the modern algorithms and it proved superior in all unimodal test functions. DA is less performed in multi-modal and composite test functions compared to MBA. Also, the obtained results were compared using the Wilcoxon rank-sum test to confirm their statistical importance. Moreover, to present an MBA's ability on real-world applications, it applied to minimize the time of completing several jobs by several persons. Our case study was a call center that involves a group of specific caller's profiles that should be assigned to a group of agents' profiles to minimize the average call time between them.

In future works, we will implement and apply multi-objective optimization problems on the MBA, and also MBA can be used to solve different optimization applications, such as

a) Assigning machines to factory orders or origins to inbound doors and the outputs to outbound doors in the warehouse management systems [37].
b) Assigning aircraft to a specific gate in the airports, where the right decision-making considers one of the important problems that face airports [38].
c) Assigning sales /marketing people to sales territories.
d) Assigning contracts to bidders by systematic bid evaluation.
e) Designing more efficient train scheduling systems on a bi-directional train strategy [39].

Meanwhile, for more performance evaluation, the MBA can be compared with other optimization algorithms, such as Dwarf Mongoose Optimization Algorithm [40], Ebola Optimization Search Algorithm [41], Reptile Search Algorithm [42], Arithmetic Optimization Algorithm [43] and Aquila Optimizer[44]. In addition, the MBA can be utilized with other optimization algorithms for possible future works suggestions, such as LSTM-ALO [45], RVM-IMRFO [46], ANFIS-GBO [47], ELM-PSOGWO [48], LSSVM-IMVO [49], SVR-SAMOA [50], ANN-EMPA [51] and ELM-CRFOA [52]. For future reading, the authors advise the reader could optionally read the following research works [53], [54], [55], [56], [57], [58], [59], [60], [61], [62], [63], [64], [65], [66], [67], [68], [69], [70].

**Acknowledgments**

Different universities provided the facilities and ongoing assistance necessary to carry out this work, which the authors sincerely appreciate.

**Compliance with Ethical Standards**

**Conflict of interest:** None.

**Funding:** Not received.